\documentclass[journal,twoside]{ieeecolor}

\usepackage{arxiv}
\usepackage{cite}
\usepackage{amsmath,amssymb,amsfonts}
\usepackage{algorithmic}
\usepackage{graphicx}
\usepackage{textcomp}
\usepackage{bbm} 
\usepackage{booktabs}
\usepackage{multirow}
\usepackage{pifont}
\usepackage[hidelinks]{hyperref}
\usepackage[table,dvipsnames]{xcolor}
\usepackage{multirow}
\usepackage{makecell}
\usepackage{amssymb}
\usepackage[table]{xcolor}
\usepackage{colortbl}

\makeatletter
\def\ps@titlepagestyle{%
  \def\@oddhead{\hfill\thepage}%
  \def\@evenhead{\thepage\hfill}%
  \let\@oddfoot\@empty
  \let\@evenfoot\@empty
}
\makeatother
\begin{document}

\title{Surg-UniWorld: A Unified Surgical World Model with Multimodal Control Experts}

\author{%
Rulin Zhou,
Wanhao Liu,
Guoheng Ma,
Liangjin Shao,
Qiujie Song,
Yidu Wang,
Guankun Wang, 
Tong Chen,
Long Bai,
Luping Zhou,
and Hongliang Ren,
\thanks{R. Zhou, W. Liu, G. Ma, L. Shao, Q. Song, Y. Wang, G. Wang, L. Bai and H. Ren are with the Department of Electronic Engineering, The Chinese University of Hong Kong, Hong Kong SAR 999077, China and Shenzhen Loop Area Institute, Shen Zhen 518055, China (email: zhourulin@link.cuhk.edu.hk; liuwanhao@mails.gdut.edu.cn; 12211611@mail.sustech.edu.cn; leonking-shaw@link.cuhk.edu.hk; songqiujie@link.cuhk.edu.hk; 1155233119@link.cuhk.edu.hk; gkwang@link.cuhk.edu.hk; b.long@ieee.org; hlren@ee.cuhk.edu.hk).}
\thanks{T. Chen and L. Zhou are from the School of Electrical and Computer Engineering, Faculty of Engineering, the University of Sydney, Sydney, NSW 2006, Australia (email: tong.chen1@sydney.edu.au; luping.zhou@sydney.edu.au).}
}

\maketitle

\begin{abstract}
Controllable surgical world models can provide a generative foundation for surgical artificial intelligence and simulation by synthesizing realistic instrument--tissue interactions. However, existing methods lack a unified multimodal control paradigm, while direct fusion of heterogeneous visual conditions often causes anatomical distortion, instrument appearance drift, and temporally inconsistent interactions. In this work, we propose {Surg-UniWorld}, a unified surgical world model with multimodal control experts. Surg-UniWorld first constructs a {Hierarchical Surgical Anchor} from first-frame appearance and hierarchical semantic masks to preserve persistent scene identity, anatomical organization, and interaction boundaries. {Anchor-Relative Modality Experts} then interpret edge, depth, and optical-flow evidence relative to the shared anchor, capturing complementary boundary, geometric, and motion information. A {Multimodal Control Expert} further performs contribution-preserving stage-wise composition of the activated modality increments and generates control hints for the Wan2.2 video diffusion backbone. To support multimodal surgical world modeling, we further construct Cholec80-SurgWAM, a benchmark for controllable surgical video generation. Extensive experiments demonstrate that Surg-UniWorld consistently outperforms existing controllable video generation methods and surgical world-model baselines in generation quality, temporal consistency, and multimodal controllability. Code and video demonstrations are available at~\url{https://surg-uniworld.pages.dev/}.

\end{abstract}


\begin{figure}[t]
\centering
\centerline{\includegraphics[width=\linewidth]{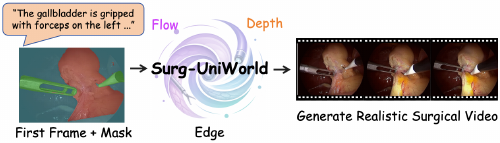}}
    \caption{Overview of Surg-UniWorld for multimodally controlled generation of
coherent instrument--tissue interaction videos. The framework supports plug-and-play integration of arbitrary combinations of edge, depth, and optical-flow controls.}
    \label{fig:overview}
    \vskip -0.2in
\end{figure}

\section{Introduction}
\label{sec:introduction}

\IEEEPARstart{V}{ideo-based} world models learn scene appearance, motion, and temporal dynamics from large-scale video data, enabling prediction of future states from current observations and potential actions~\cite{brooks2024video, qin2024worldsimbench}. Surgical world models extend this capability to operative environments, supporting robotic automation, procedural understanding, and high-fidelity
surgical video generation~\cite{koju2025surgical, he2026cosmoshsurgicallearningsurgicalrobot}. Within this paradigm, controllable surgical video generation imposes explicit semantic, geometric, and motion constraints on the generated scene. Such controllability is particularly valuable for expanding datasets with rare surgical events and underrepresented instrument--tissue interactions, generating targeted rollouts for robotic policy learning, and simulating dynamic endoscopic environments for surgical training. However, general-purpose world models are not tailored to laparoscopic scenes, where anatomical organization, instrument identity, tissue deformation, and instrument--tissue interactions must remain consistent across space and time~\cite{scheikl2022sim}. The central challenge is therefore to coordinate heterogeneous control signals while preserving anatomical stability, appearance consistency, and coherent instrument--tissue dynamics~\cite{chen2025far}.


These advances have also been extended to surgical video generation and surgical world modeling. Endora and SurgSora establish realistic and controllable surgical video synthesis, while HieraSurg and Surgical Vision World Model further move toward structured future-state prediction through semantic scene evolution and latent action modeling~\cite{li2024endora, chen2025surgsora, biagini2025hierasurg, koju2025surgical}. More recent approaches, including SAW and Cosmos-H-Surgical, explicitly formulate video generation as surgical world modeling for controllable instrument--tissue interaction synthesis and robotic policy learning~\cite{rapuri2026saw, he2026cosmoshsurgicallearningsurgicalrobot, ali2025world}. Despite this progress, existing methods rely on task-specific controls or fixed modality combinations, lacking a unified framework that distinguishes and integrates heterogeneous surgical cues by their roles in scene modeling.

A central limitation of existing approaches is that heterogeneous control signals are often treated as equivalent inputs, although they play fundamentally different roles in the modeling of surgical scenes. The first-frame appearance and hierarchical semantic masks define relatively persistent scene identities and anatomical layouts, and should therefore serve as stable world anchors. In contrast, edge, depth, and optical flow provide complementary evidence about boundaries, geometry, and motion, but their availability and reliability may vary across scenes and time. Directly concatenating or jointly injecting these signals into a shared control stream can introduce redundant or conflicting guidance, resulting in anatomical distortion, appearance drift, and temporally inconsistent interactions. Effective surgical world modeling therefore requires an anchor-centered mechanism that separately interprets modality-specific evidence and adaptively integrates it according to its contribution to the evolving surgical scene.

To address these limitations, we propose \textbf{Surg-UniWorld}, a unified surgical world model with multimodal control experts for generating controllable instrument--tissue interaction dynamics.
As illustrated in Fig.~\ref{fig:overview}, a textual instruction specifies the intended surgical action, while the first frame and hierarchical semantic masks provide persistent observations of scene appearance and spatial organization. Surg-UniWorld first constructs a stable appearance--structure anchor to preserve instrument identity,
tissue appearance, anatomical layout, and region boundaries throughout generation. Edge, depth, and optical flow are subsequently interpreted by dedicated modality experts to capture complementary boundary,
geometric, and motion evidence. The resulting anchor-relative increments are then combined in a contribution-preserving manner to produce stage-wise control hints for the pretrained Wan2.2 video diffusion backbone. This anchor-centered expert formulation enables compositional control over scene structure, visual appearance, and interaction dynamics while reducing anatomical distortion, appearance drift, and temporal inconsistency.

The main contributions are summarized as follows:

\begin{itemize}
   \item \textbf{Hierarchical Surgical Anchor.}
    First-frame appearance and hierarchical semantic masks are organized into a shared surgical anchor that preserves
    instrument and tissue identities, anatomical organization, and interaction boundaries throughout generation.
    
    \item \textbf{Anchor-Relative Modality Experts.}
    Dedicated edge, depth, and optical-flow experts interpret heterogeneous control evidence relative to the same
    Hierarchical Surgical Anchor, supporting individual modalities and arbitrary modality subsets.
    
    \item \textbf{Multimodal Control Expert.}
    It performs contribution-preserving stage-wise composition of the activated anchor-relative increments and generates control hints for the Wan2.2 video diffusion backbone.

    \item \textbf{Multimodal surgical world-modeling benchmark.}
    We construct Cholec80-SurgWAM with hierarchical masks and multimodal controls. We further establish a comprehensive benchmark for evaluating controllable surgical world models across different combinations of modalities.
\end{itemize}

\section{Related Work}

\subsection{Video World Models and Multimodal Control}

Recent advances in video foundation models have provided scalable generative backbones for learning complex spatiotemporal dynamics. LTX-Video combines a highly compressed Video-VAE with transformer-based latent diffusion, enabling efficient video generation within a compact spatiotemporal latent space~\cite{hacohen2024ltx}. Wan further scales the diffusion-transformer paradigm through large-scale video pretraining and supports diverse text-, image-, and editing-conditioned generation tasks~\cite{wan2025wan}. Moving beyond conventional visual synthesis, Cosmos-Predict2 formulates video generation as Video2World simulation, predicting plausible future visual states from textual and visual contexts for downstream Physical AI applications~\cite{he2026cosmoshsurgicallearningsurgicalrobot}. Although these models provide strong generative priors for modeling dynamic scenes, fine-grained control over scene appearance, structure, geometry, and motion still requires explicit conditioning mechanisms.

Controllable diffusion addresses this limitation by incorporating explicit structural and multimodal conditions into pretrained generative backbones. ControlNet establishes the basic paradigm of injecting spatial signals, such as edges, depth, segmentation, and pose, through trainable conditional branches~\cite{zhang2023adding}. VACE introduces a pluggable Context Adapter for LTX-Video- and WAN-based backbones, organizing reference, mask, generation, and editing conditions through a unified video-conditioning interface~\cite{jiang2025vace}. Similarly, Cosmos-Transfer extends world generation to multiple spatial controls, including segmentation, depth, and edges, and adaptively weights their contributions across spatial regions~\cite{ali2025world}. 
Nevertheless, these methods predominantly organize heterogeneous modalities as generic control streams in a shared feature space. They do not explicitly distinguish persistent appearance and semantic structure from optional boundary, geometry, and motion evidence, nor organize their complementary roles around hierarchical semantic regions.

\subsection{Surgical Video Generation and Surgical World Models}
Surgical video generation synthesizes temporally coherent endoscopic or laparoscopic sequences from visual, semantic, or action conditions. It must preserve tool--tissue structure and motion despite occlusion, specular reflections, and tissue deformation. TPG-VAE separately models content, motion, and surgical gesture priors
for future-frame prediction in robot-assisted surgery~\cite{gao2021future}. Endora developed a dedicated spatiotemporal architecture for endoscopic video generation~\cite{li2024endora}, while SurGen and Ophora explored text-guided generation for laparoscopic and ophthalmic surgical videos~\cite{cho2024surgen,li2025ophora}. Recent methods have therefore incorporated increasingly structured surgical conditions. SurgSora combines object-specific RGB-D features, segmentation cues, multi-scale optical flow, and user-defined trajectories for motion-controllable generation~\cite{chen2025surgsora}. HieraSurg instead adopts a two-stage formulation that first predicts future semantic maps from surgical phases and then renders videos using action and panoptic information~\cite{biagini2025hierasurg}. 

More recently, surgical world models have sought to capture action-dependent scene evolution for interactive simulation and robotic learning. Surgical Vision World Model infers latent actions from unannotated surgical videos and uses them to support action-controllable video generation~\cite{koju2025surgical}. SAW conditions surgical action synthesis on a reference frame, a tissue affordance mask, tool-action language, and two-dimensional tool-tip trajectories, while employing depth consistency during training to encourage geometrically plausible tool--tissue interactions~\cite{rapuri2026saw}. At a broader scale, the Cosmos world-model family provides complementary capabilities for physical simulation. Cosmos-Predict models future visual states from textual, image, or video context, whereas Cosmos-Transfer supports controllable world transformation using spatial conditions such as segmentation, depth, and edges~\cite{he2026cosmoshsurgicallearningsurgicalrobot}. Cosmos-H-Surgical-Simulator directly generates surgical manipulation videos from da~Vinci master-hand control signals, enabling action-conditioned simulation of robotic surgical procedures~\cite{he2026cosmoshsurgicallearningsurgicalrobot}. Despite this progress, existing surgical generation methods do not explicitly couple hierarchical instrument and tissue regions with region-specific appearance memories, limiting consistent and fine-grained control from edge, depth, and flow conditions.

\begin{figure}[t]
\centering
\centerline{\includegraphics[width=\linewidth]{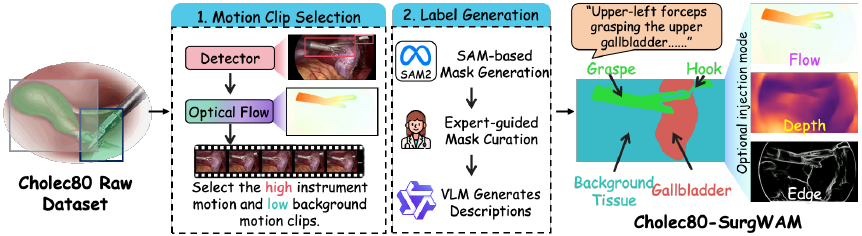}}
    \caption{Construction pipeline of Cholec80-SurgWAM, including motion-guided clip selection, expert-curated hierarchical mask generation, surgical description annotation, and optional multimodal controls.}
    \label{fig:dataset_pipeline}
    \vskip -0.05in
\end{figure}

\begin{table}[t]
\centering
\caption{Statistics of the Cholec80-SurgWAM dataset.}
\label{tab:dataset_statistics}
\begin{tabular}{lccc}
\toprule
Split & Clips & Sampled Frames & Masks \\
\midrule
Train (video01--60) & 5,104 & 250,096 & 484,820 \\
Test (video71--80)  &   897 &  43,953 &  88,901 \\
\midrule
Total               & 6,001 & 294,049 & 573,721 \\
\bottomrule
\end{tabular}
\end{table}

\section{Cholec80-SurgWAM Dataset}

We construct Cholec80-SurgWAM from the public Cholec80 dataset following the pipeline in Fig.~\ref{fig:dataset_pipeline}. Instruments are first localized, and WAFT optical flow is used to select clips with pronounced instrument motion and limited background motion, reducing the influence of global camera movement~\cite{wang2025waft}. SAM2 then provides initial masks, which are manually reviewed and refined into hierarchical annotations of instruments, target tissues, and background tissues~\cite{ravi2025sam}. A vision--language model generates descriptions of the corresponding instrument--tissue interactions. Aligned depth, optical-flow, and edge controls are obtained using Depth Anything 3, WAFT, and HED, respectively~\cite{lin2025depth,wang2025waft,xie2015holistically}.

Cholec80-SurgWAM contains 6,001 49-frame laparoscopic clips, including 5,104 training clips from video01--60 and 897 test clips from video71--80, as summarized in Table~\ref{tab:dataset_statistics}. Overall, it comprises 294,049 sampled frames and 573,721 curated masks, together with aligned textual, depth, optical-flow, and edge annotations for evaluating controllable surgical video generation under flexible modality configurations.

\section{Method}

\begin{figure*}[t]
\centering
\centerline{\includegraphics[width=0.85\linewidth]{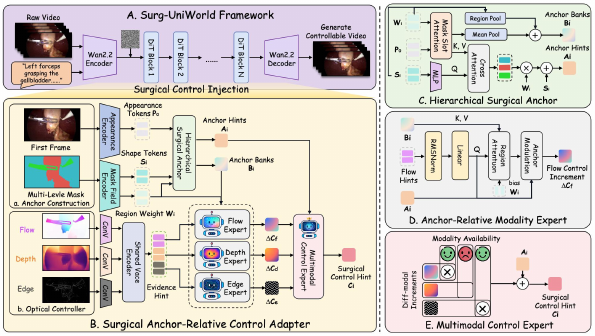}}
    \caption{Overview of Surg-UniWorld and the proposed Surgical Anchor-Relative Control Adapter (Surg-ARCA).
The Hierarchical Surgical Anchor establishes persistent appearance and structural references. Optional edge, depth,
and optical-flow conditions are interpreted by Anchor-Relative Modality Experts, and the Multimodal Control
Expert combines their anchor-relative increments into stage-wise surgical control hints for the pretrained Wan2.2 backbone.}
    \label{fig:method}
    \vskip -0.2in
\end{figure*}

\subsection{Problem Formulation and Framework Overview}
\label{sec:framework_overview}

Surg-UniWorld consists of a pretrained Wan2.2 video diffusion backbone and the proposed Surgical Anchor-Relative Control Adapter (Surg-ARCA). Let the target surgical video and the generated video be denoted by
$V=\{I_t\}_{t=0}^{T-1}$ and
$\hat{V}=\{I_0,\hat{I}_1,\ldots,\hat{I}_{T-1}\}$,
respectively. Here, $I_0$ is the given reference frame, which provides the initial visual context for subsequent generation. It is preserved throughout the denoising process, while the model synthesizes the remaining $T-1$ frames. The corresponding video-level surgical description is denoted by $Y$.

The temporally aligned hierarchical semantic masks are represented as
$\mathcal{M}=\{M_t\}_{t=0}^{T-1},\qquad
M_t=\{M_t^{\mathrm{inst}},M_t^{\mathrm{fg}},M_t^{\mathrm{bg}}\}.
$
where the three components denote the instrument, foreground-tissue, and background-tissue regions, respectively. In addition to the always-available inputs $\{I_0,y,\mathcal{M}\}$, the model can optionally receive control sequences from different modalities, including edge, depth, and optical flow:
$
\mathcal{E}=\{E_t\}_{t=0}^{T-1},\qquad
\mathcal{D}=\{D_t\}_{t=0}^{T-1},\qquad
\mathcal{F}=\{F_t\}_{t=0}^{T-1}.
$

Let $\mathcal{U}=\{\mathcal{E},\mathcal{D},\mathcal{F}\}$ and let $\mathcal{U}_S$ denote any available subset of these controls. The controllable surgical video generation task is formulated as
\begin{equation}
\hat{V}=G_{\theta}\!\left(I_0,y,\mathcal{M},\mathcal{U}_S\right),
\qquad
\mathcal{U}_S\subseteq\{\mathcal{E},\mathcal{D},\mathcal{F}\}.
\label{eq:task_definition}
\end{equation}
When $\mathcal{U}_S=\varnothing$, generation is conditioned only on the text, reference appearance, and hierarchical semantic structure. Otherwise, one or more modalities further constrain the boundary, geometry, and motion of the generated surgical video.

As illustrated in Fig.~\ref{fig:method}, Surg-ARCA produces stage-wise control hints that are injected into selected DiT blocks of the pretrained Wan2.2 backbone. Let $L=30$ denote the total number of DiT blocks and
\begin{equation}
\mathcal{L}_{\mathrm{ctrl}}
=
\{l_i\}_{i=1}^{N}
\subseteq
\{1,\ldots,L\}
\label{eq:control_layers}
\end{equation}
denote the selected control injection layers, where $N$ is the number of control stages. At the $i$-th control stage, the backbone feature is updated as
\begin{equation}
h_{l_i+1}
=
F_{l_i}
\left(
h_{l_i};t,y
\right)
+
\eta C_i,
\qquad
i=1,\ldots,N,
\label{eq:control_injection}
\end{equation}
where $F_{l_i}$ denotes the $l_i$-th Wan2.2 DiT block, $C_i$ is the stage-wise control hint produced by Surg-ARCA, and $\eta$ controls the overall injection strength. Blocks outside $\mathcal{L}_{\mathrm{ctrl}}$ follow the original Wan2.2 forward path. The construction and multimodal composition of $C_i$ are introduced in the following sections.

\subsection{Hierarchical Surgical Anchor}
\label{sec:hierarchical_anchor}

In complex laparoscopic scenes, surgical instruments, target tissues, and background tissues exhibit distinct spatial hierarchies and visual characteristics. A single semantic mask is therefore insufficient to jointly represent region identity, local boundaries, and instrument--tissue interaction relationships. Moreover, semantic masks provide structural constraints but cannot preserve instrument material, tissue texture, or scene illumination. Without a stable appearance reference, generated videos may suffer from instrument identity drift and inconsistent tissue appearance. To address these limitations, we use temporally aligned hierarchical semantic masks to define the spatial organization of the surgical scene and employ the first frame as a region-specific appearance reference. Together, they form a stable surgical anchor shared by all optional control modalities, rather than being treated as additional controls equivalent to edge, depth, or optical flow.

Given the reference frame $I_0$, the \textbf{Appearance Encoder} maps it into a set of dense visual tokens:
$P_0=f_{\mathrm{app}}(I_0)\in\mathbb{R}^{N_0\times d}.$ In implementation, the reference frame is temporally repeated to match the video-conditioning interface, while only the first temporal token plane is retained to construct $P_0$. Therefore, $P_0$ is derived exclusively from the observed frame and provides a shared appearance reference for all control stages. Importantly, $P_0$ corresponds directly to the \emph{Appearance Tokens} shown in Fig.~\ref{fig:method}; region-specific appearance pooling is performed subsequently inside the Hierarchical Surgical Anchor rather than inside the Appearance Encoder.

The semantic mask sequence consists of three mutually exclusive leaf regions:
\begin{equation}
\mathcal{M}
=
\left[
M^{\mathrm{inst}},
M^{\mathrm{fg}},
M^{\mathrm{bg}}
\right],
\qquad
M^{\mathrm{inst}}
+
M^{\mathrm{fg}}
+
M^{\mathrm{bg}}
=1.
\label{eq:semantic_regions}
\end{equation}
At the $i$-th control stage, the masks are aligned with the corresponding spatiotemporal token grid to obtain the region weights $W_i$. Meanwhile, the masks, their spatial boundaries, and their temporal variations are jointly encoded by the \textbf{Mask Field Encoder} into shape tokens $S_i$:
\begin{equation}
W_i=\mathcal{D}_i(\mathcal{M}),
\qquad
S_i=
f_{\mathrm{mask}}^{i}
\left(
\left[
\mathcal{M},
\mathcal{B}(\mathcal{M}),
|\Delta_{\tau}\mathcal{M}|
\right]
\right).
\label{eq:shape_region_tokens}
\end{equation}
Here, $\mathcal{D}_i(\cdot)$ aligns the semantic masks with the token resolution of the $i$-th control stage, $\mathcal{B}(\cdot)$ extracts region boundaries, and $\Delta_{\tau}$ denotes temporal mask differences. We use $\tau$ for video time to distinguish it from the diffusion timestep. The region weights $W_i$ identify the locations of the instrument, foreground tissue, and background tissue, whereas the shape tokens $S_i$ encode their contours and temporal structural evolution.

As illustrated in Fig.~\ref{fig:method}, the \textbf{Hierarchical Surgical Anchor (HSA)} receives the appearance tokens $P_0$, shape tokens $S_i$, and region weights $W_i$, and produces two complementary representations: a dense anchor hint $A_i$ and a compact region anchor bank $B_i$. The appearance tokens are first organized into a region-aware key--value memory through masked attention:
\begin{equation}
\left\{
K_i^{r},V_i^{r}
\right\}_{r\in\mathcal{R}_{\mathrm{leaf}}}
=
\operatorname{MaskedAttn}_{i}
\left(
P_0;W_i
\right),
\label{eq:masked_appearance_memory}
\end{equation}
where
$\mathcal{R}_{\mathrm{leaf}}
=\{\mathrm{inst},\mathrm{fg},\mathrm{bg}\}$.
The masked attention uses the corresponding region weights as spatial biases and employs multiple learnable queries for each region. This allows the resulting memory to preserve diverse local appearance patterns, including instrument reflections, heterogeneous tissue textures, and spatially varying illumination. When $W_i$ is applied to $P_0$, only its first temporal plane, aligned with the appearance-token resolution, is used.

To construct the dense anchor hint, the shape tokens are projected into structural queries and retrieve appearance information from each region-specific memory:
\begin{equation}
\begin{aligned}
R_i^{r}
&=
\operatorname{CrossAttn}
\left(
\operatorname{MLP}_{i}(S_i),
K_i^{r},
V_i^{r}
\right),\\
A_i
&=
S_i+
\operatorname{Proj}_{i}
\left(
\sum_{r\in\mathcal{R}_{\mathrm{leaf}}}
W_i^{r}\odot R_i^{r}
\right).
\end{aligned}
\label{eq:dense_anchor_construction}
\end{equation}
This region-weighted residual fusion preserves the structural information encoded by $S_i$ while introducing appearance cues only into their semantically corresponding regions. Consequently, $A_i$ provides a dense structure--appearance representation aligned with the spatiotemporal token grid.

In parallel, HSA summarizes the region-aware appearance memory into a compact anchor bank:
\begin{equation}
B_i
=
\operatorname{RegionPool}
\left(
K_i,V_i
\right)
+
\operatorname{MeanPool}
\left(
K_i,V_i
\right),
\label{eq:region_anchor_bank}
\end{equation}
where $K_i=[K_i^{r}]_{r}$ and $V_i=[V_i^{r}]_{r}$. Region pooling preserves the identities and local appearance characteristics of individual surgical regions, whereas mean pooling supplies complementary global scene context. The dense anchor hint $A_i$ is used for token-aligned anchor modulation and multimodal fusion, while the compact anchor bank $B_i$ provides the region-aware key--value reference for the subsequent model experts.

\subsection{Anchor-Relative Modality Experts}
\label{sec:modality_experts}

Although edge, depth, and optical flow are all dense visual conditions, they encode distinct properties of a surgical scene. Edge emphasizes contours and interaction boundaries, depth describes spatial geometry, and flow captures instrument and tissue motion. Direct concatenation would mix their heterogeneous statistics and couple the model to a fixed set of inputs. We instead treat them as optional evidence and introduce a modality-specific expert $\mathcal{E}_m$ for each
$m\in\mathcal{M}_{\mathrm{opt}}
=\{\mathrm{edge},\mathrm{depth},\mathrm{flow}\}$.
The experts share the same architecture but have independent parameters. Each expert is reused across all eight control stages and interprets its modality relative to the same surgical anchor rather than redefining the scene independently.

Each available control sequence $U_m$ is first converted into a VACE-compatible latent using its validity mask $\Omega_m$. A lightweight typed condition stem then preserves its modality-specific local characteristics:
\begin{equation}
\begin{aligned}
V_m
&=
\mathcal{E}_{\mathrm{cond}}(U_m,\Omega_m),\\
\widetilde V_m
&=
V_m+
\sigma(g_m)\,
W_m^{\mathrm{out}}
\operatorname{SiLU}
\left(
W_m^{\mathrm{local}} * V_m
\right).
\end{aligned}
\label{eq:typed_condition_encoding}
\end{equation}
Here, $W_m^{\mathrm{local}}$ is a $1\times3\times3$ spatial convolution for edge and depth and a $3\times3\times3$ spatiotemporal convolution for flow. The zero-initialized $W_m^{\mathrm{out}}$ makes the stem initially preserve $V_m$ before learning modality-specific residuals. No cross-modal interaction is performed at this stage.

Let $\mathcal{L}_{\mathrm{ctrl}}=\{l_i\}_{i=1}^{N}$ denote the selected control injection layers. Each typed condition is independently processed by the shared VACE patch embedding and the corresponding context blocks:
\begin{equation}
\left\{
H_{m,i}
\right\}_{i=1}^{N}
=
\mathcal{E}_{\mathrm{VACE}}
\left(
\operatorname{PE}(\widetilde V_m)+e_m;
x_t,t,y
\right),
\label{eq:shared_vace_encoding}
\end{equation}
where $e_m$ is a learnable modality embedding and $H_{m,i}$ is the control hint associated with layer $l_i$. All modalities share the VACE encoder parameters but use separate forward passes, without token concatenation or cross-modal attention. Missing modalities are skipped. The shared encoder provides a common feature space, while the typed stems and modality embeddings preserve modality identity. Each $H_{m,i}$ is then passed only to its corresponding expert $\mathcal{E}_m$.

At each control stage, the modality expert interprets $H_{m,i}$ relative to the surgical anchor, since the meaning and reliability of a local control cue depend on its semantic region. Optical flow describes instrument motion in instrument regions but tissue deformation in foreground regions. Depth characterizes instrument--tissue ordering and tissue surface geometry, while edge cues emphasize instrument contours and contact boundaries rather than irrelevant texture or illumination changes. Directly injecting these features could therefore propagate ambiguous evidence across unrelated regions. We use the compact anchor bank $B_i$ to provide region-specific references and the region weights $W_i$ to guide each token toward its corresponding anchor prototypes.

After RMS normalization and linear projection, the modality hint is mapped into a query:
\begin{equation}
Q_{m,i}
=
W_m^{q}
\operatorname{RMSNorm}
\left(
H_{m,i}
\right).
\label{eq:modality_expert_query}
\end{equation}
Meanwhile, the anchor bank is projected into the key and value features
$(K_i^{B},V_i^{B})=\phi_{kv}(B_i)$.
The region-grounded modality context is then computed as
\begin{equation}
Z_{m,i}^{\mathrm{reg}}
=
\operatorname{softmax}
\left(
\frac{
Q_{m,i}(K_i^{B})^{\top}
}{
\sqrt{d_a}
}
+
\eta_m\log(W_i+\epsilon)
\right)
V_i^{B}.
\label{eq:region_biased_expert_attention}
\end{equation}
Here, $W_i$ is broadcast over the prototypes associated with each semantic region. The resulting bias encourages tokens to retrieve appearance and structural references from their corresponding surgical regions, reducing interference between instruments, tissues, and background.

Although $B_i$ provides compact regional references, it does not retain dense token-level structure. We therefore further modulate the grounded modality context using the dense anchor hint $A_i$:
\begin{equation}
\Delta C_{m,i}
=
\operatorname{AnchorMod}_{m}
\left(
Z_{m,i}^{\mathrm{reg}},
A_i
\right).
\label{eq:anchor_relative_increment}
\end{equation}
This modulation converts the optional control into an increment relative to the stable surgical anchor, rather than allowing it to overwrite the scene structure directly. Consequently, $\Delta C_{m,i}$ preserves the semantic layout and appearance established by HSA while introducing only the modality-specific information supported by $H_{m,i}$.

\subsection{Multimodal Control Expert}
\label{sec:multimodal_composition}

To integrate the Hierarchical Surgical Anchor with optional edge, depth, and optical-flow evidence before injection into the Wan2.2 DiT backbone, we introduce the Multimodal Control Expert (MCE). At each control stage, the MCE performs contribution-preserving composition by combining the dense anchor hint $A_i$ with the anchor-relative control increments $\Delta C_{m,i}$ produced by the available Anchor-Relative Modality Experts, yielding the final surgical control hint $C_i$.

Let
$\mathcal{M}_{c}
=
\{\mathrm{edge},\mathrm{depth},\mathrm{flow}\}$
and $S\subseteq\mathcal{M}_{c}$ denote the set of available controls. The availability indicator $a_m=\mathbb{I}[m\in S]$ is determined directly from the input; unavailable modalities are skipped during VACE encoding and expert inference. Each expert increment is scaled by a learnable stage-wise factor
$\lambda_{m,i}=\sigma(s_{m,i})\in(0,1)$,
which captures the varying importance of modality $m$ across the control hierarchy without input-dependent prediction or cross-modal normalization.

The surgical control hint at stage $i$ is composed as
\begin{equation}
C_i(S)
=
A_i
+
\sum_{m\in\mathcal{M}_{c}}
a_m\lambda_{m,i}\Delta C_{m,i}.
\label{eq:contribution_preserving_composition}
\end{equation}
When no optional modality is available, the model reduces to anchor-only generation with $C_i(\varnothing)=A_i$. Because the expert increments are added without joint transformation or normalization, activating or removing one modality changes only its own contribution and leaves the remaining controls unchanged. The resulting $C_i(S)$ is injected into the corresponding Wan2.2 DiT block $l_i$ following Eq.~\eqref{eq:control_injection}.

\subsection{Training Objectives}
\label{sec:training_objectives}
We follow the standard latent flow-matching objective. Let
$S\subseteq\mathcal{M}_{c}$ denote the set of available optional controls, and let
$\hat v_S=v_{\theta}(z_t,t,y,I_0,\mathcal{M},\mathcal{U}_S)$
denote the predicted velocity, with $v^{\star}$ being its flow-matching target. We define:
\begin{equation}
\|E\|_{Q\Omega,p}^{p}
=
\frac{
\sum Q\odot\Omega\odot |E|^{p}
}{
\sum Q\odot\Omega
},
\label{eq:weighted_error}
\end{equation}
where $\Omega$ is the surgical importance field defined below. Since the first frame is provided as the reference condition, its latent temporal plane is excluded through the validity mask $Q$.

\subsubsection{Region-aware flow-matching loss $\mathcal{L}_{\mathrm{FM}}$}
A uniform flow-matching objective can be dominated by large background regions, although instrument regions and interaction boundaries are more critical for surgical video generation. We therefore construct a surgical importance field:
\begin{equation}
\Omega
=
1+
\sum_{r\in\mathcal{R}_{\mathrm{leaf}}}
\alpha_r M^r
+
\mathcal{B}(M^{\mathrm{inst}})
+
|\Delta_{\tau}M^{\mathrm{inst}}|,
\label{eq:surgical_importance_field}
\end{equation}
where
$\mathcal{R}_{\mathrm{leaf}}
=\{\mathrm{inst},\mathrm{fg},\mathrm{bg}\}$.
We set
$\alpha_{\mathrm{inst}}=2.0$,
$\alpha_{\mathrm{fg}}=1.0$, and
$\alpha_{\mathrm{bg}}=0.5$ for all experiments. The boundary and temporal terms further emphasize instrument contours and moving interaction regions. The resulting region-aware flow-matching objective is
\begin{equation}
\mathcal{L}_{\mathrm{FM}}
=
\|\hat v_S-v^{\star}\|_{Q\Omega,2}^{2}.
\label{eq:region_aware_fm_loss}
\end{equation}
This formulation integrates region and boundary supervision directly into flow matching without requiring an additional latent edge loss.

\subsubsection{Temporal Structure Loss $\mathcal{L}_{\mathrm{TS}}$}
Although $\mathcal{L}_{\mathrm{FM}}$ improves reconstruction in important regions, it does not explicitly constrain inter-frame variations. To suppress instrument flickering, contour discontinuities, and inconsistent tissue motion, we match the temporal velocity differences:
\begin{equation}
\mathcal{L}_{\mathrm{TS}}
=
\left\|
\Delta_{\tau}\hat v_S
-
\Delta_{\tau}v^{\star}
\right\|_{\bar Q\bar\Omega,2}^{2},
\label{eq:temporal_structure_loss}
\end{equation}
where
$\bar\Omega_{\tau}
=(\Omega_{\tau}+\Omega_{\tau-1})/2$
and $\bar Q$ is the corresponding valid mask for adjacent latent frames. The weighting focuses temporal supervision on instruments and their surrounding interaction regions.

\subsubsection{Control Benefit Loss $\mathcal{L}_{\mathrm{CB}}$}
The zero-initialized expert projections stabilize optimization but may cause an expert to remain close to an inactive branch. We therefore evaluate the effect of removing an active modality $m\in S$. Let
$\mathcal{R}_S=\|\hat v_S-v^{\star}\|_{Q\Omega,2}^{2}$
denote the prediction risk under control set $S$. The loss is
\begin{equation}
\mathcal{L}_{\mathrm{CB}}
=
\max\left(
0,\,
\mathcal{R}_S
-
(1-\rho)
\operatorname{sg}
\bigl(\mathcal{R}_{S\setminus\{m\}}\bigr)
\right),
\label{eq:control_benefit_loss}
\end{equation}
where $\operatorname{sg}(\cdot)$ denotes stop-gradient and $\rho$ is a relative improvement margin. This objective encourages every activated expert to provide a measurable reduction in denoising error.

\subsubsection{Marginal Consistency Loss $\mathcal{L}_{\mathrm{MC}}$}
Although the control hints are additively composed, the nonlinear DiT backbone may alter an expert's effect when other controls are present. We therefore encourage the marginal contribution of modality $m$ to remain consistent between single- and multi-modal settings:
\begin{equation}
\mathcal{L}_{\mathrm{MC}}
=
\left\|
\left(
\hat v_S-\hat v_{S\setminus\{m\}}
\right)
-
\left(
\hat v_{\{m\}}-\hat v_{\varnothing}
\right)
\right\|_{Q\Omega,1}.
\label{eq:marginal_consistency_loss}
\end{equation}
This constraint supports predictable control composition without requiring cross-modal normalization or an additional fusion network.

\subsubsection{Overall objective $\mathcal{L}$}
To reduce the computational cost of counterfactual inference, we
evaluate the control benefit and marginal consistency losses every
$K=16$ optimization steps. Let
$\chi_k=\mathbb{I}[k\bmod K=0]$, where $k$ is the current optimization
step. The complete training objective is
\begin{equation}
\mathcal{L}
=
\mathcal{L}_{\mathrm{FM}}
+
\lambda_{\mathrm{TS}}\mathcal{L}_{\mathrm{TS}}
+
\chi_k
\left(
\lambda_{\mathrm{CB}}\mathcal{L}_{\mathrm{CB}}
+
\lambda_{\mathrm{MC}}\mathcal{L}_{\mathrm{MC}}
\right).
\label{eq:overall_training_objective}
\end{equation}
We set
$\lambda_{\mathrm{TS}}=0.1$ and
$\lambda_{\mathrm{CB}}=\lambda_{\mathrm{MC}}=0.01$
in all experiments. Here, $S$ denotes the set of available optional
controls, excluding the always-on surgical anchor. For anchor-only
samples ($S=\varnothing$), only $\mathcal{L}_{\mathrm{FM}}$ and
$\mathcal{L}_{\mathrm{TS}}$ are applied. The control benefit loss
$\mathcal{L}_{\mathrm{CB}}$ is activated when $|S|\geq1$, while the
marginal consistency loss $\mathcal{L}_{\mathrm{MC}}$ is additionally activated when $|S|\geq2$.

\begin{table*}[t]
\centering
\caption{
Quantitative comparison with existing video generation and control methods.
$\uparrow$ indicates that higher is better, while $\downarrow$ indicates that
lower is better.
}
\label{tab:main_comparison}

\setlength{\tabcolsep}{3.5pt}
\renewcommand{\arraystretch}{1.08}

\resizebox{0.8\textwidth}{!}{%
\begin{tabular}{llcccccccc}
\toprule

\textbf{Model}
& \textbf{Control Configuration}
& \textbf{PSNR} $\uparrow$
& \textbf{SSIM} $\uparrow$
& \textbf{LPIPS} $\downarrow$
& \textbf{FVD} $\downarrow$
& \textbf{FID} $\downarrow$
& \textbf{Edge F1} $\uparrow$
& \textbf{Depth si-RMSE} $\downarrow$
& \textbf{Flow EPE} $\downarrow$
\\

\midrule

LTXV-2B~\cite{hacohen2024ltx}
& --
& 14.996
& 0.546
& 0.480
& 1229.801
& 71.627
& 0.108
& 0.206
& 3.911
\\

Wan2.2-5B~\cite{wan2025wan}
& --
& 17.044
& 0.632
& 0.327
& 288.250
& 14.835
& 0.159
& 0.138
& 3.197
\\

Cosmos-H-Predict-2B~\cite{he2026cosmoshsurgicallearningsurgicalrobot}
& --
& 16.326
& 0.606
& 0.340
& 202.483
& 11.329
& 0.146
& 0.126
& 2.724
\\

SurgSora~\cite{chen2025surgsora}
& --
& 15.341
& 0.524
& 0.298
& 286.440
& 30.739
& 0.234
& 0.174
& 3.048
\\

Endora~\cite{li2024endora}
& --
& 11.000
& 0.325
& 0.773
& 3391.060
& 118.599
& 0.045
& 0.277
& 12.188
\\

\midrule

\multirow{4}{*}{Cosmos-H-Transfer-2B~\cite{he2026cosmoshsurgicallearningsurgicalrobot}}
& Mask
& 13.613
& 0.434
& 0.522
& 344.761
& 61.851
& 0.338
& 0.154
& 2.711
\\

& Depth
& 12.127
& 0.370
& 0.663
& 331.035
& 59.827
& 0.252
& 0.209
& 3.342
\\

& Edge
& 16.304
& 0.456
& 0.337
& 584.788
& 57.092
& {0.347}
& 0.134
& 2.124
\\

& Flow
& 14.329
& 0.308
& 0.595
& 492.314
& 99.534
& 0.317
& 0.211
& 1.947
\\

\midrule

\multirow{4}{*}{VACE-Wan2.2-5B~\cite{jiang2025vace}}
& Mask
& 18.119
& 0.644
& 0.274
& 104.461
& 11.312
& 0.428
& 0.112
& 2.335
\\

& Depth
& 17.368
& 0.635
& 0.302
& 111.888
& 11.668
& 0.345
& 0.117
& 3.236
\\

& Edge
& 17.103
& 0.622
& 0.294
& 125.737
& 12.350
& 0.444
& 0.110
& 3.197
\\

& Flow
& 18.103
& 0.642
& 0.267
& 131.049
& 12.608
& 0.329
& 0.112
& 2.297
\\

\midrule

\multirow{4}{*}{ControlNet-Wan2.2-5B~\cite{zhang2023adding}}
& Mask
& 17.441
& 0.654
& 0.286
& 349.332
& 18.878
& 0.330
& 0.124
& 2.094
\\

& Depth
& 17.992
& 0.657
& 0.300
& 290.766
& 14.579
& 0.294
& 0.129
& 2.974
\\

& Edge
& 18.499
& 0.677
& 0.209
& 252.322
& 14.524
& 0.328
& 0.101
& 2.312
\\

& Flow
& 18.018
& 0.646
& 0.305
& 312.219
& 19.685
& 0.309
& 0.133
& 3.160
\\

\midrule

\multirow{8}{*}{\textbf{Surg-UniWorld}}
& Mask
& 18.804
& 0.657
& 0.297
& 270.996
& 10.019
& 0.301
& 0.137
& 3.567
\\

& Mask+Depth
& 18.951
& 0.669
& 0.289
& 258.298
& 8.936
& 0.321
& 0.102
& 2.959
\\

& Mask+Edge
& 20.159
& 0.702
& 0.207
& 100.654
& 7.960
& 0.459
& 0.107
& 1.981
\\

& Mask+Flow
& 19.946
& 0.698
& 0.225
& 142.344
& 7.585
& 0.440
& 0.122
& 1.742
\\

& Mask+Edge+Depth
& 20.648
& 0.708
& 0.197
& 118.882
& 7.731
& 0.433
& 0.099
& 1.480
\\

& Mask+Edge+Flow
& \underline{20.833}
& \underline{0.716}
& \underline{0.191}
& \underline{94.243}
& 7.659
& 0.461
& \underline{0.098}
& \textbf{1.308}
\\

& Mask+Depth+Flow
& 20.231
& 0.702
& 0.206
& 107.686
& \underline{7.576}
& \underline{0.469}
& 0.099
& 1.901
\\

& All
& \textbf{21.250}
& \textbf{0.722}
& \textbf{0.186}
& \textbf{92.981}
& \textbf{7.359}
& \textbf{0.506}
& \textbf{0.095}
& \underline{1.327}
\\

\bottomrule
\end{tabular}%
}

\vspace{4pt}

\end{table*}

\begin{figure*}[t]
\centering
\centerline{\includegraphics[width=0.85\linewidth]{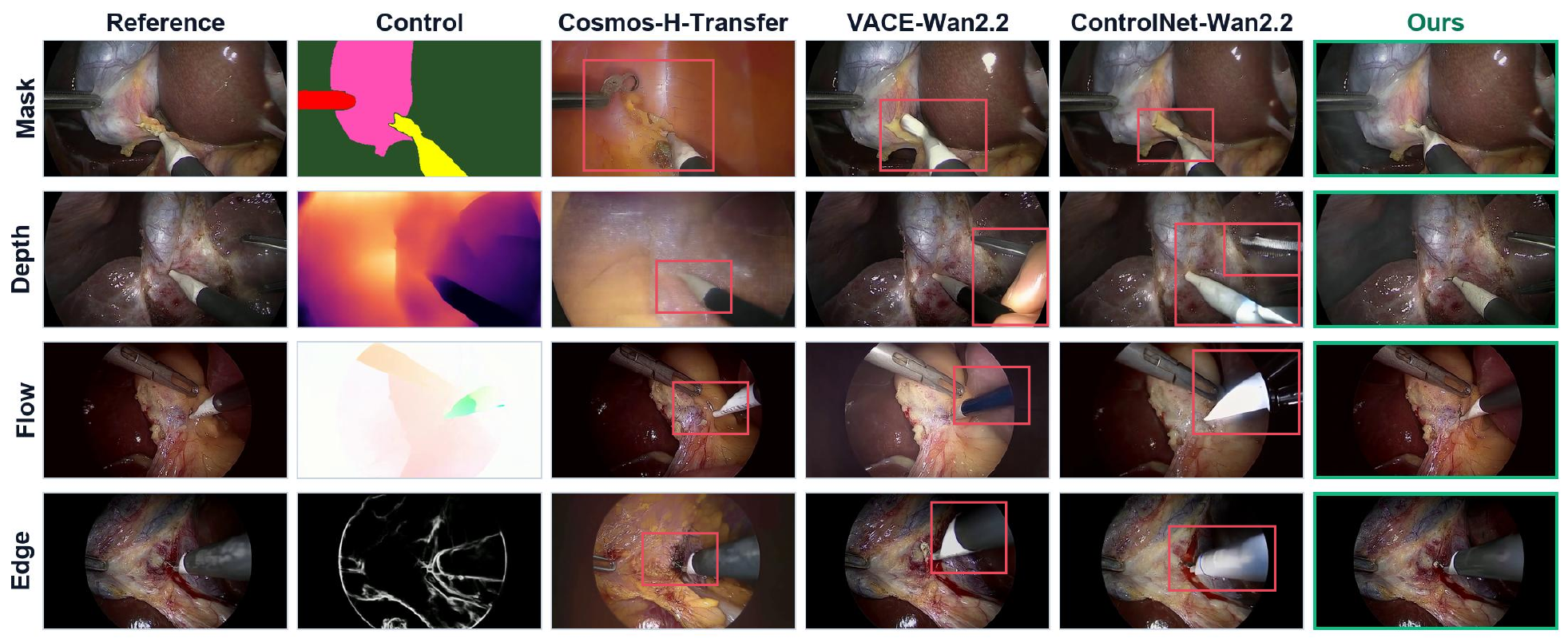}}
    \caption{
Qualitative comparison of controllable surgical video generation.
Representative results are shown under different conditions. Surg-UniWorld better preserves instrument appearance, instrument--tissue boundaries, anatomical structure, and motion consistency while following the corresponding controls. }
    \label{fig:qualitative_comparison}
    \vskip -0.15in
\end{figure*}

\section{Experiments}

\subsection{Experimental Setup and Details}
\label{sec:implementation_details}

\subsubsection{Dataset and Baselines}
All experiments are conducted on Cholec80-SurgWAM using 49-frame clips at a resolution of $832\times480$. We consider two evaluation settings: surgical video prediction from text and the first frame, covering LTXV-2B~\cite{hacohen2024ltx}, Wan2.2-5B~\cite{wan2025wan}, Cosmos-H-Predict-2B~\cite{he2026cosmoshsurgicallearningsurgicalrobot}, SurgSora~\cite{chen2025surgsora}, and Endora~\cite{li2024endora}; and controllable generation with additional spatial or motion conditions, covering Cosmos-H-Transfer-2B~\cite{he2026cosmoshsurgicallearningsurgicalrobot}, VACE-Wan2.2~\cite{jiang2025vace}, and ControlNet-Wan2.2~\cite{zhang2023adding}. Each baseline is evaluated with the conditions supported by its interface, whereas Surg-UniWorld uses first-frame appearance and hierarchical masks as the surgical anchor and supports arbitrary subsets of optional depth, edge, and optical-flow controls.

\subsubsection{Evaluation Metrics}

Following the unified video-generation evaluation protocol adopted in VACE~\cite{jiang2025vace}, we evaluate the surgical videos generated using PSNR, SSIM, LPIPS, FID and FVD. PSNR and SSIM measure pixel-level fidelity and structural similarity, whereas LPIPS evaluates perceptual similarity. FID measures frame-level distribution quality, while FVD captures spatiotemporal realism. Higher PSNR and SSIM and lower LPIPS, FID, and FVD indicate better performance. To evaluate control adherence, we also report Edge F1, Depth si-RMSE, and Flow EPE for configurations containing the corresponding control modality. Following prior controllable generation protocols~\cite{alhaija2025cosmos}, each condition is re-extracted from the generated video using a fixed modality estimator and compared with the input control. Higher Edge F1 indicates better boundary alignment, whereas lower Depth si-RMSE and Flow EPE indicate better geometric and motion adherence, respectively. All metrics are further evaluated within the instrument, foreground-tissue, and background-tissue regions using the temporally aligned semantic masks.


\subsubsection{Implementation Details}
For all modality-conditioned methods, the pretrained generative backbone is frozen and only the additional control modules are optimized for 7,500 steps on four NVIDIA H100 GPUs using AdamW with a learning rate of $2\times10^{-5}$. The number of control injection layers is fixed to $N=8$ for Surg-UniWorld, VACE-Wan2.2, and ControlNet-Wan2.2. To support arbitrary
control availability, the active modality subset $S\subseteq\mathcal{M}_c$ is sampled according to
\begin{equation}
p(S)=
\begin{cases}
0.20, & S=\varnothing\ \text{or}\ S=\mathcal{M}_c,\\
0.10, & \text{otherwise},
\end{cases}
\end{equation}
where $\mathcal{M}_c=\{\mathrm{edge},\mathrm{depth}, \mathrm{flow}\}$.  This strategy emphasizes the anchor-only and
full-control settings while retaining balanced exposure to all
single- and dual-modal combinations.

\begin{figure*}[ht]
\centering
\centerline{\includegraphics[width=0.9\linewidth]{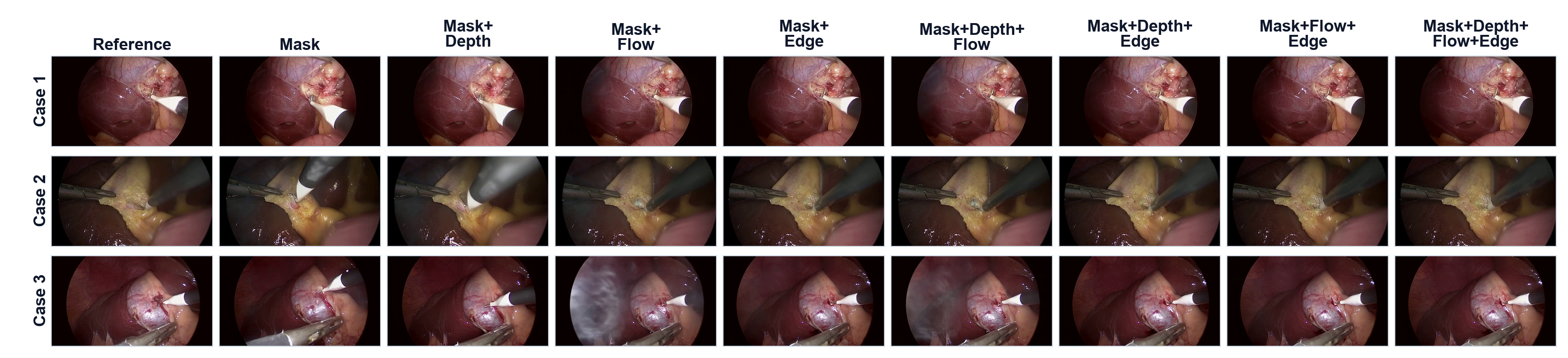}}
    \caption{
Qualitative analysis of control-modality composition in
Surg-UniWorld. Multimodal combinations better preserve
appearance, structure, interaction boundaries, and motion
coherence than single-control settings. }
    \label{fig:multimodal_ours}
    \vskip -0.1in
\end{figure*}

\begin{figure*}[ht]
\centering
\centerline{\includegraphics[width=0.8\linewidth]{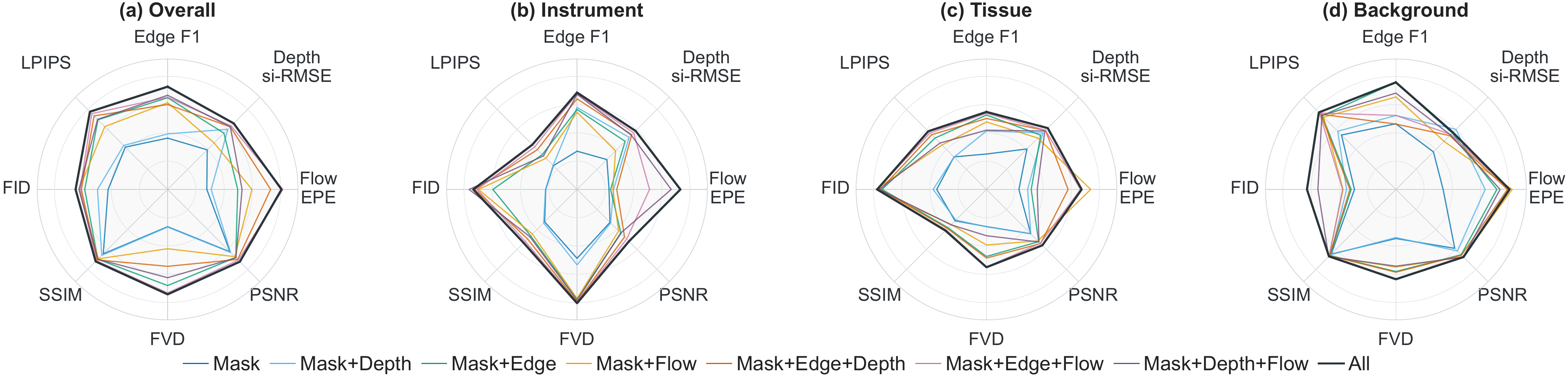}}
    \caption{
Effect of control-modality composition on overall and
region-specific performance. Radar plots summarize normalized generation-quality and
control-adherence scores for
(a) the full frame,
(b) instrument regions,
(c) foreground-tissue regions, and
(d) background-tissue regions
across all optional-control combinations.
Lower-is-better metrics are direction-aligned such that a larger
radius consistently indicates better performance.}
    \label{fig:radar}
    \vskip -0.15in
\end{figure*}

\subsection{Comparison with Baselines}

\subsubsection{Quantitative Evaluation}

As shown in Table~\ref{tab:main_comparison}, the full-control configuration of Surg-UniWorld achieves the strongest overall performance, with a PSNR of 21.250, an SSIM of 0.722, an LPIPS of 0.186, an FVD of 92.981, and an FID of 7.359. Compared with the strongest baseline results, it improves PSNR by 2.751\,dB and reduces FVD and FID by approximately 11.0\% and 34.9\%,
respectively. The consistent improvements across pixel-level, perceptual, and spatiotemporal metrics indicate that the gains extend beyond frame reconstruction to overall temporal realism. The full-control configuration also obtains an Edge F1 of 0.506 and a Depth SI-RMSE of 0.095, while the Mask+Edge+Flow configuration achieves the lowest Flow EPE of 1.308. These results demonstrate that the proposed anchor-centered multimodal control framework improves generation quality while maintaining strong adherence to heterogeneous surgical conditions.

\subsubsection{Qualitative Evaluation}

As shown in Fig.~\ref{fig:qualitative_comparison}, we compare different controllable generators under mask, depth, optical-flow, and edge conditions. Existing methods generally reproduce the coarse content of the reference scene but still exhibit noticeable inconsistencies in appearance and structure. Cosmos-H-Transfer often deviates from the reference tissue appearance and anatomical layout, whereas VACE-Wan2.2 and ControlNet-Wan2.2 better preserve the global scene but remain susceptible to instrument deformation, boundary misalignment, and local tissue distortion. As highlighted by the red boxes, these methods may also introduce implausible content unrelated to the surgical scene or produce locally corrupted regions. In contrast, Surg-UniWorld more consistently preserves instrument identity, tissue appearance, anatomical structure, and instrument--tissue spatial relationships while adhering to the corresponding controls. Specifically, mask conditioning maintains the prescribed semantic layout and object contours, depth conditioning improves the geometric organization of foreground and background regions, optical-flow conditioning produces more coherent instrument and tissue motion, and edge conditioning better retains instrument boundaries and fine structural details.

\begin{figure}[t]
\centering
\centerline{\includegraphics[width=0.9\linewidth]{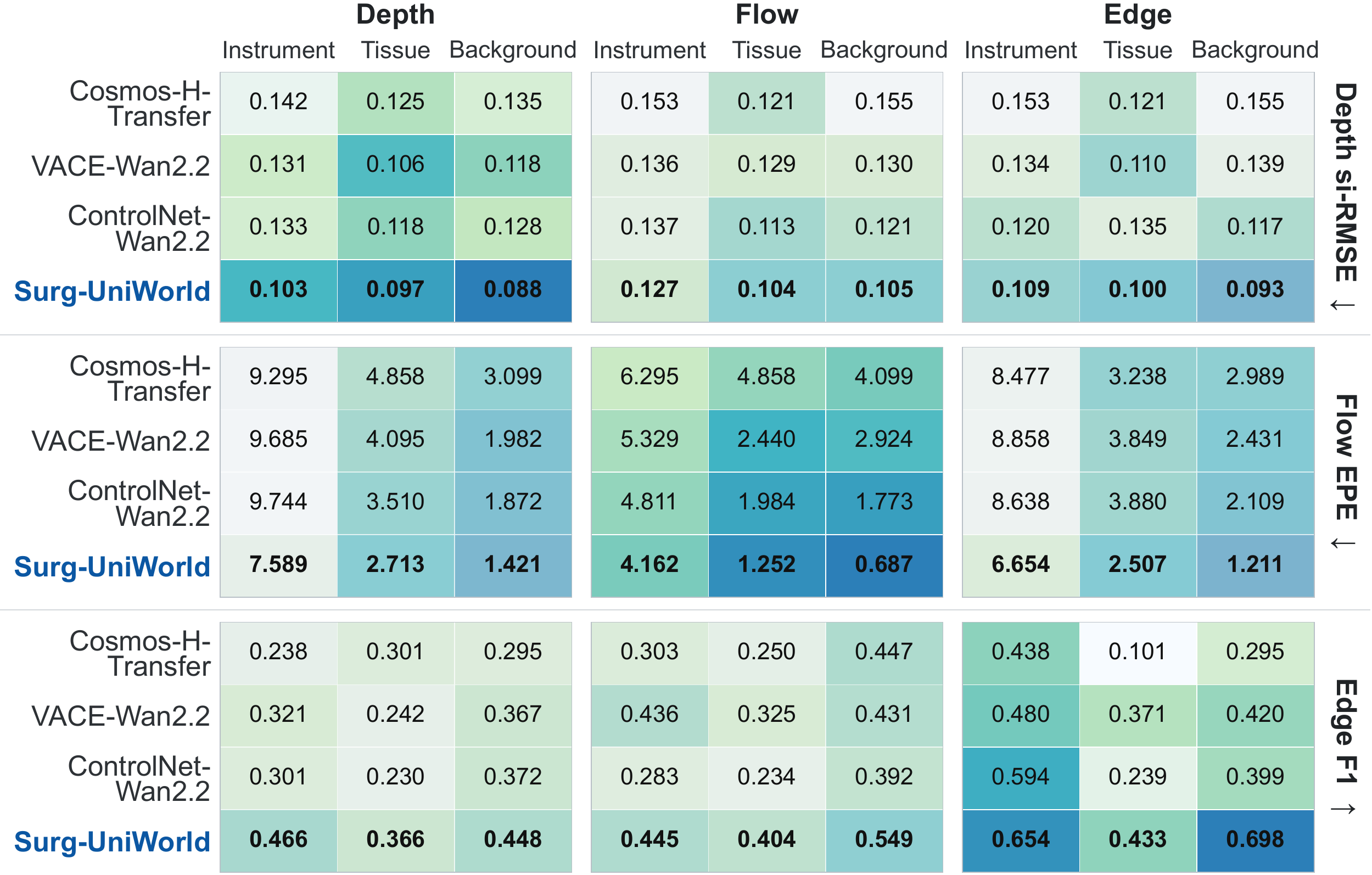}}
    \caption{Region-wise comparison of multimodal control adherence.
Depth SI-RMSE, Flow EPE, and Edge F1 are evaluated separately
within instrument, foreground-tissue, and background-tissue regions.}
    \label{fig:cross_model_heatmap}
    \vskip -0.25in
\end{figure}

\subsection{Analysis of Multimodal Control}

\subsubsection{Region-wise Modality Specificity and Control Adherence}

Fig.~\ref{fig:cross_model_heatmap} presents a cross-modality evaluation of depth, optical-flow, and edge controls within the instrument, foreground-tissue, and background-tissue regions. The modality-matched entries consistently exhibit the strongest performance, indicating that each injected modality primarily improves its corresponding control property. Specifically, depth conditioning achieves Depth SI-RMSE values of 0.103, 0.097, and 0.088 across the three regions; flow conditioning obtains Flow EPE values of 4.162, 1.252, and 0.687; and edge conditioning reaches Edge F1 scores of 0.654, 0.433, and 0.698. Moreover, Surg-UniWorld achieves the best matched control-adherence result across all nine modality--region pairs compared with the baselines. This diagonal dominance demonstrates that the Anchor-Relative Modality Experts preserve distinct geometric, motion, and boundary information while enabling effective region-aware control.

\definecolor{lossbg}{RGB}{224,236,248}
\definecolor{samplingbg}{RGB}{229,241,232}
\definecolor{groundingbg}{RGB}{250,235,218}

\begin{table*}[t]
    \centering

    \caption{
    Ablation studies on training objectives,
    control-subset sampling, and the effects of region grounding
    and stage-wise scaling.
    }
    \label{tab:combined_ablation}

    \begingroup
    \small
    \setlength{\tabcolsep}{4.2pt}
    \renewcommand{\arraystretch}{1.10}

    \resizebox{0.8\textwidth}{!}{%
    \begin{tabular}{
        ll
        !{\vrule width 0.5pt}
        cccccccc
    }
        \toprule

        \multicolumn{2}{c!{\vrule width 0.5pt}}{
            \textbf{Configuration}
        }
        & \textbf{PSNR}$\uparrow$
        & \textbf{SSIM}$\uparrow$
        & \textbf{LPIPS}$\downarrow$
        & \textbf{FVD}$\downarrow$
        & \textbf{FID}$\downarrow$
        & \textbf{Edge F1}$\uparrow$
        & \textbf{Depth SI-RMSE}$\downarrow$
        & \textbf{Flow EPE}$\downarrow$
        \\

        \midrule

        \rowcolor{lossbg}
        \multicolumn{2}{l!{\vrule width 0.5pt}}{
            \textbf{Loss}
        }
        &
        \multicolumn{8}{c}{
            \textbf{Training Objective Ablation}
        }
        \\

        \multicolumn{2}{l!{\vrule width 0.5pt}}{
            w/o $\mathcal{L}_{\mathrm{TS}}$
        }
        & 21.073
        & 0.715
        & 0.192
        & 109.384
        & 7.683
        & \underline{0.501}
        & \underline{0.097}
        & 1.462
        \\

        \multicolumn{2}{l!{\vrule width 0.5pt}}{
            w/o $\mathcal{L}_{\mathrm{CB}}$
        }
        & 20.982
        & 0.714
        & 0.196
        & 102.764
        & 7.821
        & 0.487
        & 0.100
        & 1.405
        \\

        \multicolumn{2}{l!{\vrule width 0.5pt}}{
            w/o $\mathcal{L}_{\mathrm{MC}}$
        }
        & \underline{21.108}
        & \underline{0.718}
        & \underline{0.190}
        & \underline{98.716}
        & \underline{7.612}
        & 0.498
        & \underline{0.097}
        & \underline{1.369}
        \\

        \multicolumn{2}{l!{\vrule width 0.5pt}}{
            \textbf{Full objective (Ours)}
        }
        & \textbf{21.250}
        & \textbf{0.722}
        & \textbf{0.186}
        & \textbf{92.981}
        & \textbf{7.359}
        & \textbf{0.506}
        & \textbf{0.095}
        & \textbf{1.327}
        \\

        \midrule

        \rowcolor{samplingbg}
        \multicolumn{2}{l!{\vrule width 0.5pt}}{
            \textbf{Control-Subset Sampling}
        }
        &
        \multicolumn{8}{c}{
            \textbf{Control-Subset Training Strategies}
        }
        \\

        \multicolumn{2}{l!{\vrule width 0.5pt}}{
            All-controls only
        }
        & 20.582
        & 0.696
        & 0.211
        & 118.734
        & 8.462
        & 0.472
        & 0.104
        & 1.492
        \\

        \multicolumn{2}{l!{\vrule width 0.5pt}}{
            Uniform subset sampling
        }
        & \underline{21.089}
        & \underline{0.717}
        & \underline{0.192}
        & \underline{98.624}
        & \underline{7.648}
        & \underline{0.498}
        & \underline{0.098}
        & \underline{1.372}
        \\

        \multicolumn{2}{l!{\vrule width 0.5pt}}{
            Random sampling
        }
        & 21.061
        & 0.716
        & 0.193
        & 99.438
        & 7.691
        & 0.496
        & 0.099
        & 1.384
        \\

        \multicolumn{2}{l!{\vrule width 0.5pt}}{
            \textbf{Biased subset sampling (Ours)}
        }
        & \textbf{21.250}
        & \textbf{0.722}
        & \textbf{0.186}
        & \textbf{92.981}
        & \textbf{7.359}
        & \textbf{0.506}
        & \textbf{0.095}
        & \textbf{1.327}
        \\

        \midrule

        \rowcolor{groundingbg}
        \textbf{Region Grounding}
        & \textbf{Stage-wise Scale}
        &
        \multicolumn{8}{c}{
            \textbf{Region Grounding and Stage-Wise Scaling}
        }
        \\

        $\times$
        & $1$
        & 20.450
        & 0.699
        & 0.211
        & 121.839
        & 8.720
        & \underline{0.511}
        & 0.097
        & {1.446}
        \\

        $\checkmark$
        & $1$
        & \underline{20.920}
        & \underline{0.714}
        & \underline{0.195}
        & \underline{105.631}
        & \underline{7.919}
        & \textbf{0.524}
        & \underline{0.096}
        & \underline{1.354}
        \\

        $\times$
        & $\lambda_{m,i}$
        & 20.860
        & 0.711
        & 0.198
        & 107.442
        & 8.054
        & 0.494
        & 0.103
        & 1.385
        \\

        \textbf{\(\boldsymbol{\checkmark}\)}
        & \textbf{\(\boldsymbol{\lambda}_{m,i}\)}
        & \textbf{21.250}
        & \textbf{0.722}
        & \textbf{0.186}
        & \textbf{92.981}
        & \textbf{7.359}
        & 0.506
        & \textbf{0.095}
        & \textbf{1.327}
        \\

        \bottomrule
    \end{tabular}%
    }
    \vspace{-1.5mm}
    \endgroup
\end{table*}

\subsubsection{Effect of Multimodal Control Composition}

As reported in Table~\ref{tab:main_comparison}, the optional controls provide distinct and complementary benefits. Compared with the anchor-only configuration, denoted as ``Mask'' in the table, adding depth reduces Depth SI-RMSE from 0.137 to 0.102, adding optical flow reduces Flow EPE from 3.567 to 1.742, and adding edge increases Edge F1 from 0.301 to 0.459. These
results confirm that depth, flow, and edge primarily strengthen geometric, motion, and boundary constraints, respectively. Fig.~\ref{fig:radar} further visualizes the normalized performance profiles across the full frame and the instrument, foreground-tissue, and background-tissue regions. Multimodal configurations generally provide broader and more balanced performance than single-control settings, with the full configuration achieving the best result on seven of the eight metrics among all Surg-UniWorld configurations. The three representative cases in Fig.~\ref{fig:multimodal_ours} provide consistent visual evidence: single controls mainly improve their corresponding scene properties, whereas multimodal combinations better preserve instrument appearance, tissue structure, interaction boundaries, and motion coherence.

\begin{table}[t]
    \centering
    \caption{
        Ablation of the core architectural designs.
    }
    \label{tab:architecture_ablation}

    \begingroup
    \small
    \setlength{\tabcolsep}{3.6pt}
    \renewcommand{\arraystretch}{1.10}

    \resizebox{\columnwidth}{!}{%
    \begin{tabular}{lccccc}
        \toprule
        \textbf{Configuration}
        & \textbf{LPIPS$\downarrow$}
        & \textbf{FVD$\downarrow$}
        & \textbf{Edge F1$\uparrow$}
        & \textbf{Depth SI-RMSE$\downarrow$}
        & \textbf{Flow EPE$\downarrow$} \\
        \midrule

        Flat Appearance--Structure Anchor
        & 0.199 & 108.416 & 0.498 & 0.098 & 1.381 \\

        Shared Modality Expert
        & 0.196 & 104.128 & 0.490 & 0.097 & 1.403 \\

        w/o Anchor-Relative Modulation
        & \underline{0.191}
        & \underline{99.287}
        & \textbf{0.513}
        & \textbf{0.094}
        & \underline{1.355} \\

        \textbf{Full Architecture (Ours)}
        & \textbf{0.186}
        & \textbf{92.981}
        & \underline{0.506}
        & \underline{0.095}
        & \textbf{1.327} \\

        \bottomrule
    \end{tabular}%
    }

    \endgroup
    \vspace{-1.5mm}
\end{table}

\subsection{Ablation Studies}

\subsubsection{Ablation of Core Architectural Designs}
As shown in Table~\ref{tab:architecture_ablation}, replacing the Hierarchical Surgical Anchor with flat appearance--structure fusion increases LPIPS from 0.186 to 0.199 and FVD from 92.981 to 108.416, demonstrating the benefit of region-specific appearance memory and hierarchical region-aware anchoring. Using a shared modality expert reduces Edge F1 to 0.490 and increases Flow EPE to 1.403, confirming that boundary,
geometric, and motion evidence requires modality-specific interpretation. Removing anchor-relative modulation slightly improves Edge F1 and Depth SI-RMSE to 0.513 and 0.094, respectively, but increases LPIPS, FVD, and Flow EPE. This suggests that directly applying modality evidence may strengthen
local control conformity at the expense of perceptual quality and motion coherence.

\subsubsection{Ablation of Training Objectives} 
As shown in Table~\ref{tab:combined_ablation}, removing any auxiliary objective degrades overall performance. Without $\mathcal{L}_{\mathrm{TS}}$, FVD increases from 92.981 to 109.384 and Flow EPE from 1.327 to 1.462, confirming its importance for temporal consistency. Removing $\mathcal{L}_{\mathrm{CB}}$ produces the lowest Edge F1 of 0.487, indicating that explicitly encouraging each activated expert to reduce denoising error improves control effectiveness. Removing $\mathcal{L}_{\mathrm{MC}}$ also increases FVD and Flow EPE to 98.716 and 1.369, respectively, supporting its role in maintaining consistent modality contributions across control subsets. These results demonstrate the complementary roles of the three auxiliary objectives.

\subsubsection{Ablation of Control-subset Sampling}
Training with all controls only yields substantially worse performance, indicating limited robustness when one or more modalities are unavailable at inference time. Uniform and random subset sampling improve performance by exposing the model to different modality combinations, whereas the proposed biased sampling achieves the best results across all metrics. This
confirms that emphasizing the anchor-only and full-modal settings while retaining single- and dual-modal configurations provides an effective balance between stable generation and flexible multimodal control.

\subsubsection{Ablation of Region Grounding and Stage-Wise Scaling}

This ablation evaluates region-grounded attention in the modality
experts and learnable stage-wise scaling in the multimodal composer. Region grounding with a fixed scale reduces FVD from 121.839 to 105.631 and increases Edge F1 from 0.511 to 0.524, demonstrating the benefit of aligning modality evidence with region-specific anchor prototypes. Stage-wise scaling alone also improves generation quality, reducing FVD to 107.442 and Flow EPE to 1.385. Combining both designs achieves the strongest
overall performance, including an LPIPS of 0.186, an FVD of 92.981, and a Flow EPE of 1.327. Although fixed-scale grounding yields a slightly higher Edge F1, the complete configuration provides the best overall balance between generation quality and control fidelity.

\subsubsection{Ablation of Composition Strategies in the
Multimodal Control Expert}
\begin{figure}[t]
\centering
\centerline{\includegraphics[width=0.9\linewidth]{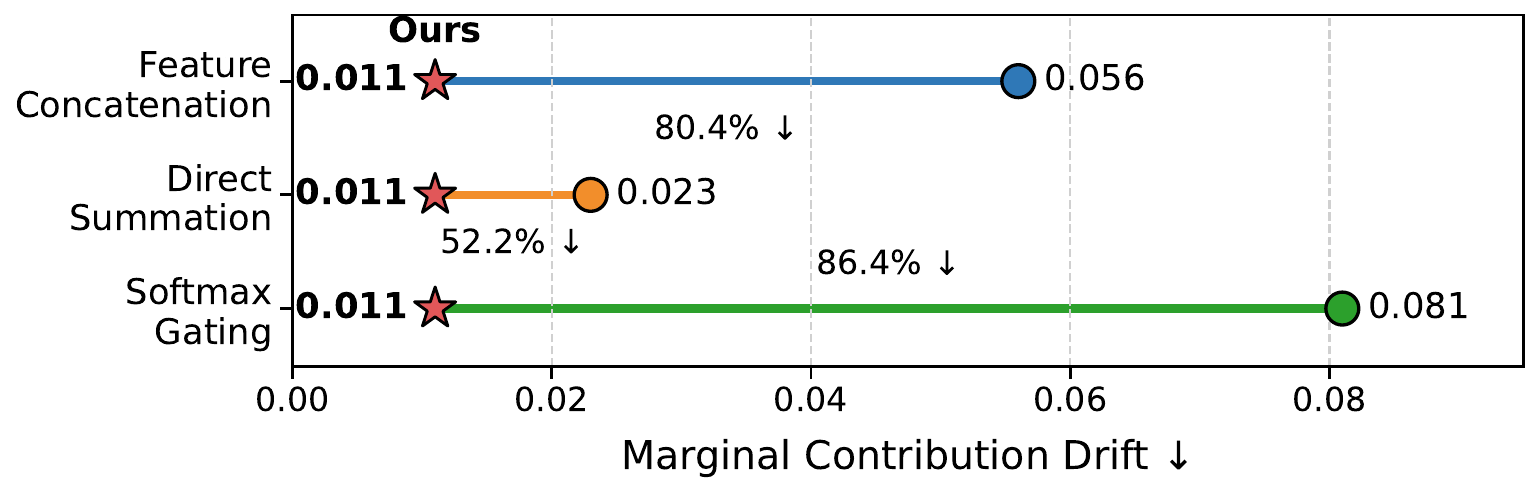}}
    \caption{Marginal contribution stability of different composition strategies. Our stage-wise composition yields the lowest drift.}
    \label{fig:marginal_contribution_stability}
    \vskip -0.2in
\end{figure}

Table~\ref{tab:fusion_strategy} and Fig.~\ref{fig:marginal_contribution_stability} compare alternative composition strategies in the Multimodal Control Expert. Direct summation achieves the strongest individual control-adherence scores, with an Edge F1 of 0.518, a Depth SI-RMSE of 0.092, and a Flow EPE of 1.281, but substantially degrades generation quality, increasing LPIPS and FVD to 0.204 and 112.476. Feature concatenation and softmax gating provide more balanced results but introduce stronger coupling among modality contributions. In contrast, the proposed contribution-preserving stage-wise composition achieves the best LPIPS of 0.186 and FVD of 92.981 while maintaining competitive boundary, geometric, and motion fidelity. It further reduces marginal contribution drift to 0.011, corresponding to reductions of 80.4\%, 52.2\%, and 86.4\% relative to feature concatenation, direct summation, and softmax gating, respectively. These results demonstrate that preserving each expert's independent contribution provides a better balance between generation quality, control fidelity, and multimodal composition stability.

\begin{table}[t]
    \centering
    \caption{
       Comparison of composition strategies in the Multimodal Control Expert.
    }
    \label{tab:fusion_strategy}

    \begingroup
    \small
    \setlength{\tabcolsep}{3.6pt}
    \renewcommand{\arraystretch}{1.10}

    \resizebox{\columnwidth}{!}{
    \begin{tabular}{
        l!{\vrule width 0.5pt}
        c c c c c
    }
        \toprule

        \textbf{Composition Strategy}
        &
        \textbf{LPIPS$\downarrow$}
        &
        \textbf{FVD$\downarrow$}
        &
        \textbf{Edge F1$\uparrow$}
        &
        \textbf{Depth SI-RMSE$\downarrow$}
        &
        \textbf{Flow EPE$\downarrow$}
        \\

        \midrule

        Feature Concatenation
        & 0.199
        & 105.842
        & 0.493
        & 0.101
        & 1.402
        \\

        Direct Summation
        & 0.204
        & 112.476
        & \textbf{0.518}
        & \textbf{0.092}
        & \textbf{1.281}
        \\

        Softmax Gating
        & \underline{0.192}
        & \underline{99.638}
        & 0.489
        & 0.098
        & 1.391
        \\

        \textbf{Stage-Wise Composition (Ours)}
        & \textbf{0.186}
        & \textbf{92.981}
        & \underline{0.506}
        & \underline{0.095}
        & \underline{1.327}
        \\

        \bottomrule
    \end{tabular}
    }
    \vspace{-1.5mm}
    \endgroup
\end{table}

\section{Conclusion}

In this work, we presented Surg-UniWorld, a unified surgical world model for controllable generation of laparoscopic instrument--tissue interactions under flexible multimodal conditions. Surg-UniWorld constructs a Hierarchical Surgical Anchor from first-frame appearance and hierarchical semantic masks, interprets edge, depth, and optical-flow evidence through Anchor-Relative Modality Experts, and combines their stage-wise increments using contribution-preserving multimodal composition. We further introduced Cholec80-SurgWAM and demonstrated through extensive comparisons and ablations that the proposed framework improves visual fidelity, temporal coherence, region-level control adherence, robustness to arbitrary modality subsets, and marginal contribution stability. These results support separating persistent surgical anchors from
optional modality evidence for scalable surgical simulation and data
augmentation. However, extreme instrument occlusion and rapid camera
motion remain challenging, motivating future work on explicit
camera-motion modeling and long-horizon action conditioning.

\bibliographystyle{IEEEtran}
\bibliography{references}

@inproceedings{gao2021future,
  title={Future frame prediction for robot-assisted surgery},
  author={Gao, Xiaojie and Jin, Yueming and Zhao, Zixu and Dou, Qi and Heng, Pheng-Ann},
  booktitle={International Conference on Information Processing in Medical Imaging},
  pages={533--544},
  year={2021},
  organization={Springer}
}

@inproceedings{li2024endora,
  title={Endora: Video generation models as endoscopy simulators},
  author={Li, Chenxin and Liu, Hengyu and Liu, Yifan and Feng, Brandon Y and Li, Wuyang and Liu, Xinyu and Chen, Zhen and Shao, Jing and Yuan, Yixuan},
  booktitle={International conference on medical image computing and computer-assisted intervention},
  pages={230--240},
  year={2024},
  organization={Springer}
}

@article{cho2024surgen,
  title={Surgen: Text-guided diffusion model for surgical video generation},
  author={Cho, Joseph and Schmidgall, Samuel and Zakka, Cyril and Mathur, Mrudang and Kaur, Dhamanpreet and Shad, Rohan and Hiesinger, William},
  journal={arXiv preprint arXiv:2408.14028},
  year={2024}
}

@inproceedings{chen2025surgsora,
  title={Surgsora: Object-aware diffusion model for controllable surgical video generation},
  author={Chen, Tong and Yang, Shuya and Wang, Junyi and Bai, Long and Ren, Hongliang and Zhou, Luping},
  booktitle={International Conference on Medical Image Computing and Computer-Assisted Intervention},
  pages={521--531},
  year={2025},
  organization={Springer}
}

@inproceedings{biagini2025hierasurg,
  title={Hierasurg: Hierarchy-aware diffusion model for surgical video generation},
  author={Biagini, Diego and Navab, Nassir and Farshad, Azade},
  booktitle={International Conference on Medical Image Computing and Computer-Assisted Intervention},
  pages={310--319},
  year={2025},
  organization={Springer}
}

@inproceedings{li2025ophora,
  title={Ophora: a large-scale data-driven text-guided ophthalmic surgical video generation model},
  author={Li, Wei and Hu, Ming and Wang, Guoan and Liu, Lihao and Zhou, Kaijing and Ning, Junzhi and Guo, Xin and Ge, Zongyuan and Gu, Lixu and He, Junjun},
  booktitle={International Conference on Medical Image Computing and Computer-Assisted Intervention},
  pages={425--435},
  year={2025},
  organization={Springer}
}

@inproceedings{koju2025surgical,
  title={Surgical vision world model},
  author={Koju, Saurabh and Bastola, Saurav and Shrestha, Prashant and Amgain, Sanskar and Shrestha, Yash Raj and Poudel, Rudra PK and Bhattarai, Binod},
  booktitle={MICCAI Workshop on Data Engineering in Medical Imaging},
  pages={1--10},
  year={2025},
  organization={Springer}
}

@article{rapuri2026saw,
  title={SAW: Toward a Surgical Action World Model via Controllable and Scalable Video Generation},
  author={Rapuri, Sampath and Seenivasan, Lalithkumar and Schneider, Dominik and Soberanis-Mukul, Roger and He, Yufan and Ding, Hao and Xu, Jiru and Yu, Chenhao and Jing, Chenyan and Guo, Pengfei and others},
  journal={arXiv preprint arXiv:2603.13024},
  year={2026}
}

@misc{he2026cosmoshsurgicallearningsurgicalrobot,
  title={Cosmos-H-Surgical: Learning Surgical Robot Policies from Videos via World Modeling},
  author={Yufan He and Pengfei Guo and Mengya Xu and Zhaoshuo Li and Andriy Myronenko and Dillan Imans and Bingjie Liu and Dongren Yang and Mingxue Gu and Yongnan Ji and Yueming Jin and Ren Zhao and Baiyong Shen and Daguang Xu},
  year={2026},
  eprint={2512.23162},
  archivePrefix={arXiv},
  primaryClass={cs.RO},
  url={https://arxiv.org/abs/2512.23162},
}

@inproceedings{zhang2023adding,
  title={Adding conditional control to text-to-image diffusion models},
  author={Zhang, Lvmin and Rao, Anyi and Agrawala, Maneesh},
  booktitle={Proceedings of the IEEE/CVF international conference on computer vision},
  pages={3836--3847},
  year={2023}
}

@inproceedings{jiang2025vace,
  title={Vace: All-in-one video creation and editing},
  author={Jiang, Zeyinzi and Han, Zhen and Mao, Chaojie and Zhang, Jingfeng and Pan, Yulin and Liu, Yu},
  booktitle={Proceedings of the IEEE/CVF International Conference on Computer Vision},
  pages={17191--17202},
  year={2025}
}

@article{wan2025wan,
  title={Wan: Open and advanced large-scale video generative models},
  author={Wan, Team and Wang, Ang and Ai, Baole and Wen, Bin and Mao, Chaojie and Xie, Chen-Wei and Chen, Di and Yu, Feiwu and Zhao, Haiming and Yang, Jianxiao and others},
  journal={arXiv preprint arXiv:2503.20314},
  year={2025}
}

@article{hacohen2024ltx,
  title={Ltx-video: Realtime video latent diffusion},
  author={HaCohen, Yoav and Chiprut, Nisan and Brazowski, Benny and Shalem, Daniel and Moshe, Dudu and Richardson, Eitan and Levin, Eran and Shiran, Guy and Zabari, Nir and Gordon, Ori and others},
  journal={arXiv preprint arXiv:2501.00103},
  year={2024}
}

@article{brooks2024video,
  title={Video generation models as world simulators},
  author={Brooks, Tim and Peebles, Bill and Holmes, Connor and DePue, Will and Guo, Yufei and Jing, Leo and Schnurr, David and Taylor, Joe and Luhman, Troy and Luhman, Eric and others},
  journal={OpenAI Blog},
  volume={1},
  number={8},
  pages={1},
  year={2024}
}

@article{qin2024worldsimbench,
  title={Worldsimbench: Towards video generation models as world simulators},
  author={Qin, Yiran and Shi, Zhelun and Yu, Jiwen and Wang, Xijun and Zhou, Enshen and Li, Lijun and Yin, Zhenfei and Liu, Xihui and Sheng, Lu and Shao, Jing and others},
  journal={arXiv preprint arXiv:2410.18072},
  year={2024}
}

@article{chen2025far,
  title={How Far Are Surgeons from Surgical World Models? A Pilot Study on Zero-shot Surgical Video Generation with Expert Assessment},
  author={Chen, Zhen and Xu, Qing and Wu, Jinlin and Yang, Biao and Zhai, Yuhao and Guo, Geng and Zhang, Jing and Ding, Yinlu and Navab, Nassir and Luo, Jiebo},
  journal={arXiv preprint arXiv:2511.01775},
  year={2025}
}

@article{scheikl2022sim,
  title={Sim-to-real transfer for visual reinforcement learning of deformable object manipulation for robot-assisted surgery},
  author={Scheikl, Paul Maria and Tagliabue, Eleonora and Gyenes, Bal{\'a}zs and Wagner, Martin and Dall'Alba, Diego and Fiorini, Paolo and Mathis-Ullrich, Franziska},
  journal={IEEE Robotics and Automation Letters},
  volume={8},
  number={2},
  pages={560--567},
  year={2022},
  publisher={IEEE}
}

@article{ali2025world,
  title={World simulation with video foundation models for physical ai},
  author={Ali, Arslan and Bai, Junjie and Bala, Maciej and Balaji, Yogesh and Blakeman, Aaron and Cai, Tiffany and Cao, Jiaxin and Cao, Tianshi and Cha, Elizabeth and Chao, Yu-Wei and others},
  journal={arXiv preprint arXiv:2511.00062},
  year={2025}
}

@article{lin2025depth,
  title={Depth anything 3: Recovering the visual space from any views},
  author={Lin, Haotong and Chen, Sili and Liew, Junhao and Chen, Donny Y and Li, Zhenyu and Shi, Guang and Feng, Jiashi and Kang, Bingyi},
  journal={arXiv preprint arXiv:2511.10647},
  year={2025}
}

@article{wang2025waft,
  title={Waft: Warping-alone field transforms for optical flow},
  author={Wang, Yihan and Deng, Jia},
  journal={arXiv preprint arXiv:2506.21526},
  year={2025}
}

@inproceedings{ravi2025sam,
  title={Sam 2: Segment anything in images and videos},
  author={Ravi, Nikhila and Gabeur, Valentin and Hu, Yuan-Ting and Hu, Ronghang and Ryali, Chaitanya and Ma, Tengyu and Khedr, Haitham and R{\"a}dle, Roman and Rolland, Chloe and Gustafson, Laura and others},
  booktitle={International Conference on Learning Representations},
  volume={2025},
  pages={28085--28128},
  year={2025}
}

@article{alhaija2025cosmos,
  title={Cosmos-transfer1: Conditional world generation with adaptive multimodal control},
  author={Alhaija, Hassan Abu and Alvarez, Jose and Bala, Maciej and Cai, Tiffany and Cao, Tianshi and Cha, Liz and Chen, Joshua and Chen, Mike and Ferroni, Francesco and Fidler, Sanja and others},
  journal={arXiv preprint arXiv:2503.14492},
  year={2025}
}

@inproceedings{xie2015holistically,
  title     = {Holistically-Nested Edge Detection},
  author    = {Xie, Saining and Tu, Zhuowen},
  booktitle = {Proceedings of the IEEE International Conference on Computer Vision},
  pages     = {1395--1403},
  year      = {2015}
}

\end{document}